\documentclass[10pt,letterpaper]{article}

\usepackage{cogsci}
\usepackage{pslatex}
\usepackage{apacite}
\usepackage{graphicx}
\usepackage{url}
\usepackage{float}
\usepackage{bbding}
\usepackage{xcolor}

\cogscifinalcopy

\newcommand{\hole}{\underline{\hspace{1em}}}

\title{Visual representation of negation: Real world data analysis on comic image design}

\author{{\large \bf Yuri Sato$^1$, Koji Mineshima$^2$, and Kazuhiro Ueda$^1$}\\
$^1$Department of General Systems Studies, 
Graduate School of Arts and Sciences, The University of Tokyo, Japan\\
satoyuri0@gmail.com, ueda@gregorio.c.u-tokyo.ac.jp\\
$^2$Department of Philosophy, Faculty of Letters, Keio University, Japan\\
minesima@abelard.flet.keio.ac.jp
}

\begin{document}

\maketitle

\begin{abstract}
There has been a widely held view that visual representations (e.g., photographs and illustrations) do not depict negation, for example, one that can be expressed by a sentence ``the train is not coming''.
This view is empirically challenged by analyzing the real-world visual representations of comic (manga) illustrations.
In the experiment using image captioning tasks, we gave people comic illustrations and asked them to explain what they could read from them. The collected data showed that 
some comic illustrations could depict negation without any aid of sequences (multiple panels) or conventional devices (special symbols).
This type of comic illustrations
was subjected to further experiments, classifying images into those containing negation and those not containing negation.
While this image classification was easy for humans, it was difficult for data-driven machines, i.e., deep learning models (CNN), to achieve the same high performance.
Given the findings, we argue that 
some comic illustrations evoke background knowledge
and thus can depict negation with purely visual elements.

\end{abstract}

\section{Introduction}
Negation plays an important role in our thinking and communication. In natural language, we can express the negation of a proposition
``the train is coming''
by making a negated sentence ``the train is not coming.'' Similarly, in symbolic logic, negation is viewed as an operator (e.g.~$\neg A$) to flip the truth value of a proposition.
The meaning and use of negation in linguistic representations
has been widely studied 
in AI and logic (Wanshing, 1996), 
semantics and pragmatics (Horn, 1989), psycholinguistics (Kaup \& Zwaan, 2003; Dale \& Duran, 2011; Nordmeyer \& Frank, 2014), and psychology of reasoning (Khemlani, Orenes, \& Johnson-Laird, 2014).

Compared to linguistic representations, it is not straightforward
to express negation in visual representations such as photographs
and illustrations.
For example, suppose that
you send a picture of a railway station platform with no trains
to let your friend know that there is no train coming.
Perhaps this visual way of communicating negation
is not as reliable as a text message.
This raises the question:
are there visual representations that 
can be recognized as expressing negation?
This is the question to be addressed in the present paper.

One influential reaction to this question
was given in the study of diagrammatic reasoning.
Among others, Barwise and Etchemendy (1992) 
have built a logic learning support system called \textit{Hyperproof}
that uses heterogeneous modules combining linguistic (symbolic) and visual representations.
Regarding the possibility of representing negation in such a hybrid system, they say,
``$\dots$ diagrams and pictures are extremely good at presenting a wealth of specific, conjunctive information. It is much harder to use them to present indefinite information, negative information, or disjunctive information '' (p.79). 
Behind this view may be found the following philosophical claim;
Wittgenstein (1914/1984) remarked in the draft leading to the \textit{Tractatus} that a picture (what is depicted) cannot be denied (Notebook, 26 Nov 1914).
A similar view is widely found in the literature on philosophy of mind and language (Heck, 2007; Crane, 2009).

Against this widely held view, we will argue that negation can be depicted in various interesting ways.
To our knowledge, empirical investigation on 
visual representations of negation 
is remarkably understudied.
While there are studies that empirically evaluate
visual representation systems
that support devices to express negation, such as Hyperproof (Stenning, Cox, \& Oberlander, 1995) and
logic diagrams (Jones, McInnes, \& Staveley, 1999; Sato \& Mineshima, 2015),
these previous studies focus on 
the type of representations that are designed and created in a top-down manner,
mainly to conduct controlled evaluation experiments.
Instead, we focus on the type of 
visual representations that 
naturally occur outside the scientific domain
and are used to express and communicate human thoughts in everyday situations.
In this sense, we take a \textit{data-driven} approach, according to which we collect and analyze \textit{real-world} visual representations that are actually used by people and survive as a design in our culture.

Among various types of real-world visual representation,
we focus on \textit{comic (\textit{manga}) illustrations} in the present study.
We first provide an analysis of syntactic components 
that make up comics.
Then we introduce a dataset that collects comic illustrations related to negation.
We introduce the image captioning task
that asks participants to explain what they can read from a given illustration.
This will show evidence that some visual representations can express negation.
To further analyze what enables illustrations to depict negation,
we introduce a task of classifying illustrations related to negation, comparing machine learning (deep learning) performance and human performance. 
Overall results suggest humans can exploit presupposed knowledge evoked by comic illustrations to recognize negation.

\begin{figure}[t]\center
\begin{minipage}{2.5cm}\footnotesize
\vspace{1em}
\hspace{1.8em}speech balloon\\

\vspace{1em}

conventional device\\
(special symbol){$\rightarrow$}
\vspace{5em}

\end{minipage}
\begin{minipage}{3.3cm}
\includegraphics[scale=0.21]{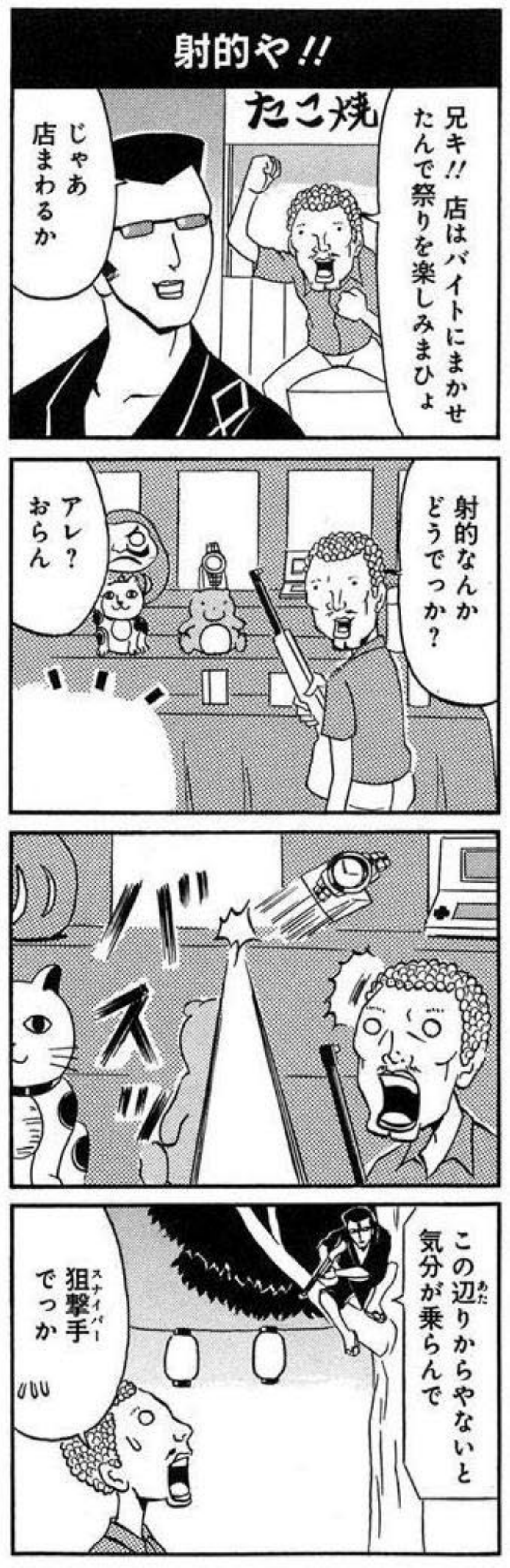}
\end{minipage}
\begin{minipage}{0.2cm}
\setlength{\unitlength}{1mm}
\begin{picture}(20,75)
\put(0,43){ \line(-1,3){10} }
\put(0,41){ \line(-1,1){10} }
\put(0,39){ \line(-1,-1){10} }
\put(0,37){ \line(-1,-3){10} }
\end{picture}
\end{minipage}
\begin{minipage}{2.2cm}\footnotesize

temporal\\ sequences\\
(multiple panels)

\end{minipage}
\caption{\footnotesize A case of comic, retrieved from p.26 of ``Tetsu-san'' {\copyright} Kengo Kawaguchi/TokyoSanseiSha}
\label{fig:tetsu}
\end{figure}
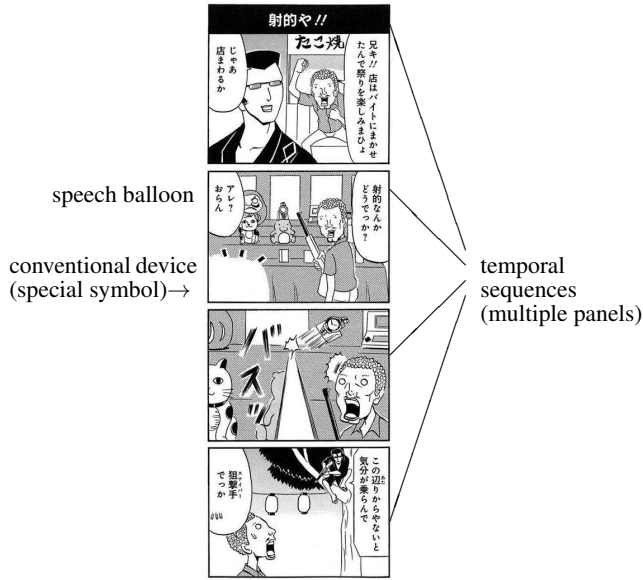

\section{Syntactic components of comics}

To understand the role of negation in visual representation, we give a preliminary analysis of syntactic components of comics,
which are best characterized in comparison with other forms of non-linguistic representations,
in particular, 
photographs and videos (cf. Sato \& Mineshima,~2020).

First, videos and comics are distinguished from photographs in that they consist of \textit{temporal sequences}, which typically represent sequences of multiple events or scenes. 
For example, in the case of the comic ``Tetsu-san'' in Fig.\ref{fig:tetsu}, the contents are presented as multiple panels in a temporal sequence.
The fact that the notion of sequence is an important element in videos and comics has been emphasized in various recent studies on images.
Thus, in the study of video captioning, 
sequential images, rather than single images, enable humans and machines to describe certain actions (e.g., jump) (e.g.,  Yeung et al., 2018).
Regarding comics,  
Iyyer et al. (2017) presented a deep learning model to predict subsequent panels according to the sequences of coherent stories.

Second, comics are distinguished from photographs and videos in that they have a variety of \textit{conventional devices} for expressing thoughts, emotions, and other non-visual properties.
These devices range from special symbols having some iconic character such as motion-lines, fight clouds,
and what Cohn~(2013) calls ``affixes'' like thundercloud and stars circling over a head  to express anger, to
more linguistic ones like speech balloons to show a character's utterance or thought (Maier, 2019).
In the comic example in Fig.\ref{fig:tetsu}, linguistic utterances are indicated inside balloons and special symbols such as effect lines are used.
By contrast, photographs and videos typically
do not involve conventional devices,
though they can be enriched with symbols or linguistic materials;  
see Giora et al. (2009), which discusses
various possible interpretations of conventional devices for negation such as crosses and lines and evaluate empirical data collected from humans.

The upshot of this analysis is that
in comics,
not only purely visual elements of illustrations but also various syntactic components play a role in delivering semantic information including negation.
Thus, to answer the question of whether negation can be depicted in visual representations, 
it is necessary to identify each potential factor contributing to
the information delivered by comics.
Accordingly, we ablate various elements from comics
by, for example, extracting a single panel (i.e., ignoring temporal sequences), hiding linguistic utterances in balloons, and removing conventional devices, and see how the resulting representations
work to depict negation.

\section{Datasets}
To study comics as complex visual representations, we introduce comic datasets collected from the real world. 

\subsection{Manga109 and masterpiece list}
We use Manga109~(Matsui et al., 2017; Aizawa et al., 2020)\footnote{The comic images included in Manga109 have been given permission to use
for research purposes. All of the comic images in this paper are from the Manga109 dataset.}, 
a standard dataset of Japanese comics.
We also use some masterpiece works not included in Manga109.
This additional set was selected from Japanese manga works 
by authors who have won two major manga artist awards; 
(1) Japan Cartoonists Association Award (JCAA), Minister of Education Award and (2) Tezuka Osamu Cultural Prize (TOCP), Special Award.
We collected award-winning works
in either JCAA,  TOCP, Shogakukan Manga Award, Kodansha Manga Award or Bungeishunju Manga Award.
Our list of masterpiece works include 
Osamu Tezuka's \textit{Black Jack}, \textit{Budda}, \textit{Hinotiri (Phoenix)}; 
Shotaro Ishinomori's \textit{Cyborg009}, \textit{Jun}, \textit{MangaNihonKeizaiNyuumon (Introduction to Manga Japan Economy)}, \textit{SabuToIchiTorimonohikae};   
Tetsuya Chiba's \textit{1$\cdot$2$\cdot$3To4$\cdot$5$\cdot$Roku}, \textit{NotariMatsutaro}, \textit{OreWaTeppei};
Chikako Mitsuhashi's \textit{ChiisanaKoiNoMonogatari};  
Shigeru Mizuki's \textit{TeleviKun (TV Kids)}.

\subsection{\textsc{NegComic} dataset}
From the Manga109 and the masterpiece list, 
we created the \textsc{NegComic} dataset in the following way.
(1)
Images related to negation that consists of multiple pages were collected by one of the authors; 122 images. 
(2)
Three evaluators (including the author in (1))
assessed whether the collected images were relevant to expressing negation.
(3)
The pages (and most relevant panels) that were judged as expressing negation by at least two evaluators were finally selected; 111 images.
In (2) and (3), 
the following instruction was given to the evaluators; 
By negation, we mean information typically paraphrased as 
``\textit{there is no \hole},''
``\textit{\hole\ does not exist},''
``\textit{\hole\ is not \hole},''
``\textit{\hole\ disappeared},''
``\textit{\hole\ is empty},''
``\textit{\hole\ cannot do \hole},''
``\textit{\hole\ does not move},''
etc.

As a baseline for data analysis, 
we also created a negation-free image set.
From the same pages as the negation images selected above,
we asked three evaluators (including one of the authors of the paper) to choose panels that are \textit{not} related to negation.
We included those panels that were selected by at least two evaluators in the negation-free image set.

We categorize negation depicted in images into two types.
(1) \textit{Existence negation} expresses absence of entities---one typically expressed by sentences such as \textit{the cup is empty},
\textit{there was not a cat here},
and \textit{I lost my wallet};
In formal notation, it corresponds to the negation of existential proposition, $\neg \exists x Px$ (there is nothing that satisfies the condition $P$).
(2) \textit{Property negation} expresses that an entity referred to does not have a property or does not perform an action.
It is typically expressed by sentences like 
\textit{the signal is not green} and \textit{my body does not move};
formally, it corresponds to formula $\neg Pc$ where $c$ refers to a particular entity and $P$ to a property (or action).
Drawing on Bloom (1970),
Nordmeyer and Frank (2014) use a similar distinction between
non-existence and truth-functional negation
to examine children's ability to comprehend negated sentences.

We annotated a gold-standard type (existence negation or property negation) to each image.
Eighty seven images were classified to existence negation and the other 24 images to property negation.
The annotations of 81 images (out of all 111 images) 
were automatically done by searching typical phrases for each type
in the comic images (we use the phrase patterns described in Table~\ref{tab:phrase} below)
and those of 26 images were manually given by two of the authors in this paper.

\section{Image captioning task}
To answer the question of whether there are visual representations that can be recognized as expressing negation, we gave people comic illustrations and asked them to explain what they could read from the illustrations. 

\subsection{Method}
\subsubsection{Participants}
Four hundred and fifty-nine participants were recruited by using an online crowdsourcing platform in Japan, CrowdWorks. 
The mean age of participants was 38.15 (SD = 9.53) with a range of 20-74 years.
All participants declared that they could read and write Japanese without difficulty (the sentences and instructions were provided in Japanese.).  
All participants gave informed consent and were paid for their participation. 
Experimental procedures were approved by the ethics committee of the University of Tokyo. 

\subsubsection{Tasks}
Regarding each image in the \textsc{NegComic} set, 
six versions were prepared:
\begin{enumerate} \setlength{\itemsep}{-0.15cm} 
    \item sequential images (the original 111 images)
    \item sequential images in the negation-free type
    \item sequential images without conventional devices 
    \item one-panel images (111 images)
    \item one-panel images in the negation-free type
    \item one-panel images without conventional devices 
\end{enumerate}

\noindent
We divide images into those with temporal sequences (sequential images) and those without (one-panel images).
Here negation-free conditions (2 and 5) served as a baseline for analysis and were compared
with sequential or one-panel images (1 vs.~2 and 4 vs.~5)
and images without conventional devices (3 vs.~2 and 6 vs.~5).
Since our focus is on the question of how pictures (illustrations) can depict negation, we ablated all linguistic materials in comic images from the \textsc{NegComic} set.
In conditions 3 and 6 (26 images), 
conventional devices which are relevant to negation were removed from the original images (conditions 1 and 4). They 
included 
broken lines tracing contours (see Fig.3b),  
radial lines (see Fig.2), 
question marks, 
crossing marks, 
blanks inside speech balloons, 
overlaid contour lines (for shaking),  
black filled panels, 
and fading lines. 
If condition 1 is better than condition 2 and condition 4 is better than condition 5 (and if condition 3 is better than condition 2 and condition 6 is better than condition 5 in case there is a conventional device), the illustration would be counted as a visual representation expressing negation.

\begin{figure}[t]\center
\begin{tabular}{c}
\includegraphics[scale=0.31]{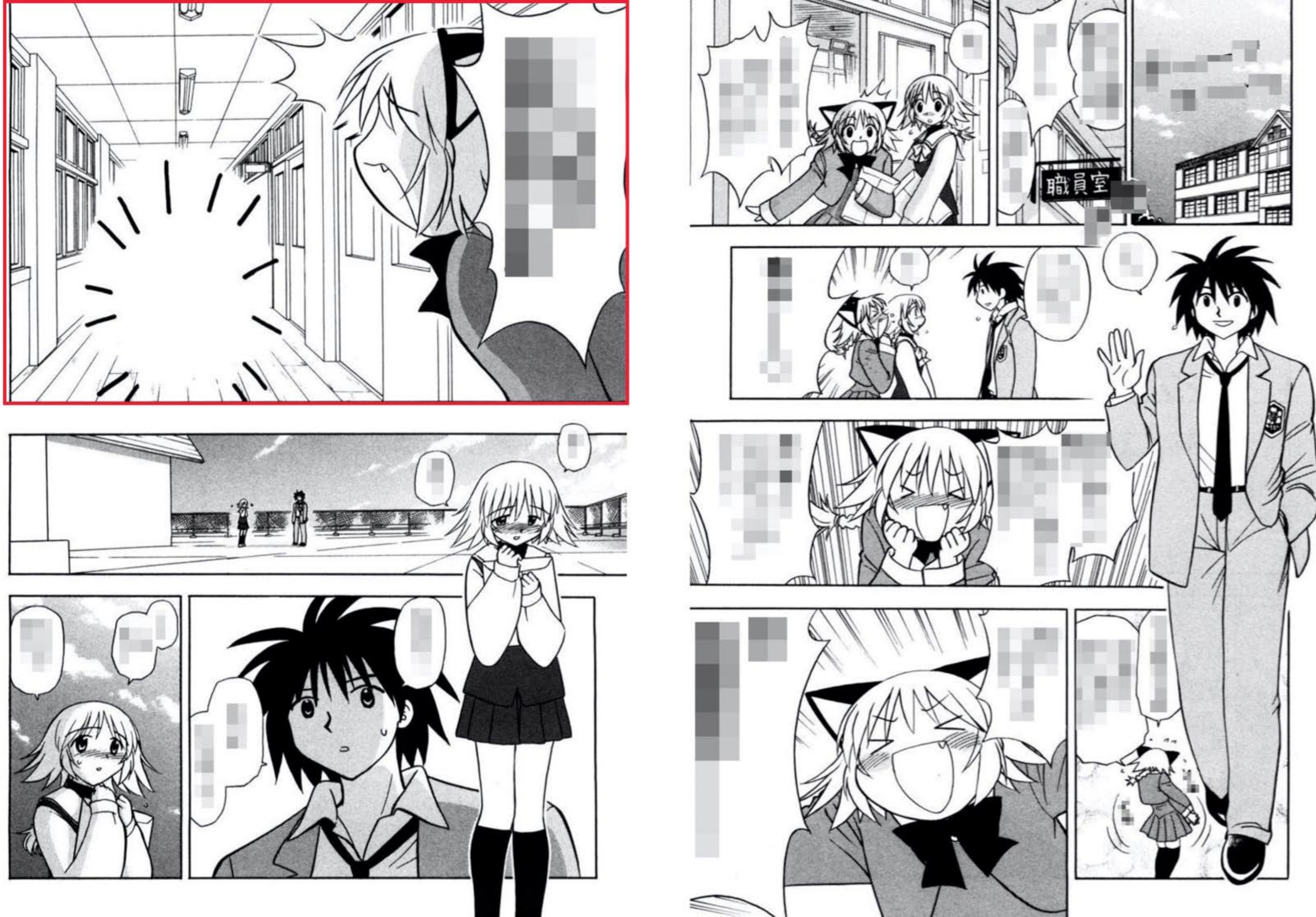}\\
(a)
\end{tabular}

\begin{tabular}{cc}
\includegraphics[scale=0.4]{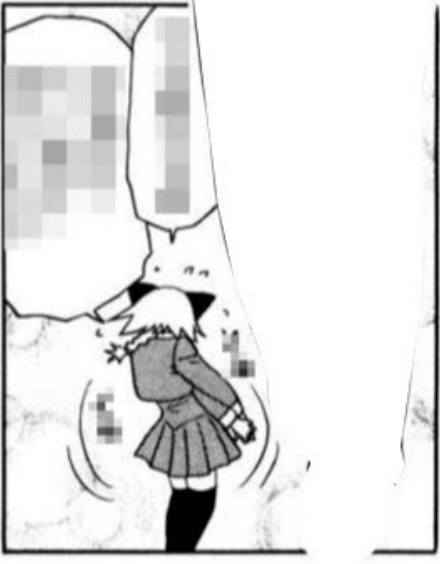}
&
\includegraphics[scale=0.355]{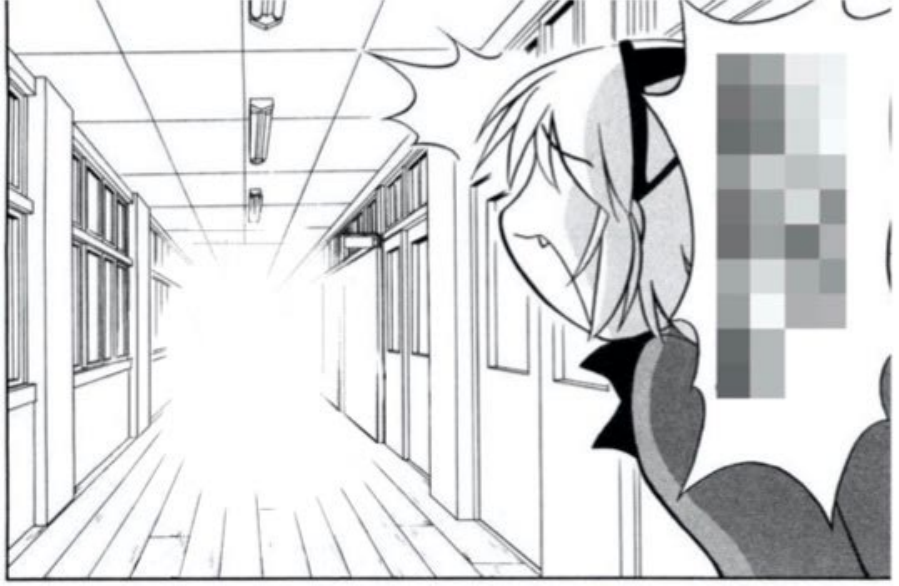}\\
(b)
&
(c)
\end{tabular}

\caption{\footnotesize
(a) Task example of condition 1 (sequential image) (the first panel of the left page, surrounded in red, is specified); 
(b) condition 5 (one-panel images in the negation-free); 
(c) condition 6 (one-panel images without conventional devices).
Retrieved from pp.21-22 of ``Arisa$^2$'' $\copyright$ Ken Yagami / Kadokawa-Shoten
}
\label{tab:pattern}
\end{figure}

The following instructions were given to the participants: 
(1)
You are given comic images (the linguistic parts are mosaicked), so please write down what you can read in Japanese.
(2) Don't write your feelings or opinions.
(3) If you are familiar with the work and clearly remember the content of the relevant part, skip this question (image).

Examples of tasks are shown in Fig.2
These images were presented with the instruction ``Explain the specified panel in as much detail as possible.''
Note that Fig.2c is the one-panel image that does not contain the conventional device of radial line. By contrast, the original image, Fig.2a, is sequential and has a conventional device in the panel with a red outer frame.

\subsubsection{Procedures}
Participants were randomly divided into one of the six conditions.
Each participant was 
presented with 26 items; that is, in conditions 1, 2, 4, 5, 26 items out of all task items were randomly presented. 
The task presentation and data collection were managed using an interface implemented in Qualtrics. 

\subsection{Results}
Average 23.95 text data per image (in each condition) was obtained from the participants;
meta-phrases such as ``I can't explain this image'' 
as one's own opinion
were not included in our analysis.
For example, 
regarding condition 1 in ``Arisa$^2$'' (Fig.2a), 
text data as shown in Table 1 was obtained (only data for three participants is shown here; MeCab\footnote{\url{https://github.com/taku910/mecab}} was used here for the morphological analysis of Japanese text). 
\begin{table}[t]\footnotesize
\caption{\footnotesize 
Data examples obtained in the condition 1 task of ``Arisa$^2$'' (Fig.2a); Checkmark means that the data is judged as expressing
(existence) negation. The bracketed text in the bottom row is written in Japanese.}
\begin{tabular}{clc}\\ 
\hline
\hline
1. & The boy has not been there. & \Checkmark\\
& (otokonoko ga i naku na tte i ta.) & \\
2. & The guy just disappeared. & \Checkmark\\
& (otoko no hito ga kie ta.) & \\
3. & A girl is surprised to find a boy she likes.\\
& (shoujo ga suki na ko wo mitsuke te bikkuri si te iru.) & \\
\hline
\hline\\
\end{tabular}
\vspace{-1.5em}
\end{table}

To count the occurrences of negation clauses in the obtained text data, we listed the clauses belonging to each type of negation: existence negation and property negation (Table 2).
In addition to the correct answer annotations, 
synonyms of the correct words (of the verb class) were set to be included in the negation clauses, before the data-collecting.
We used word2vec\footnote{We used Python library \textsf{gensim}.} (learned from the full texts of Japanese Wikipedia articles) to extract an initial set of synonyms, and 
selected the words that overlap with synonyms found in
Japanese WordNet\footnote{ \url{http://compling.hss.ntu.edu.sg/wnja/}} or WLSP\footnote{\url{https://github.com/masayu-a/WLSP}}.
Furthermore, 
after the experiment, 
we manually checked the data and judged which type of negation an expression with the negative morpheme (\textit{nai}) belongs to.
The results are shown in the italicized phrases in Table 2.

Occurrences of all phrases belonging to each type were counted for each image and compared to the baseline (conditions 2 and 5) using Fisher's Exact Test (with the Bonferroni correction).
For example, consider 
the above case of 
the condition 1 task of ``Arisa$^2$'' (Table 1; Fig.2a).
This case is annotated as the existence negation in advance, and so 
we check whether appearances of all the phrases in table 2(a) are included in the text data. 
In the examples in Table 1, 
text 1 and text 2 include phrases \textit{not been there}
and \textit{disappeared}, respectively; thus they are
judged as expressing negation.
By contrast, text 3 does not include any negation phrases,
so this case is judged as not expressing negation.

\begin{table}[t]\footnotesize
\caption{\footnotesize Negation phrase lists for (a) existence negation and (b) property negation; 
{\bf bold} means correct annotations,  
normal fonts mean their synonym,  
\textit{italic} means what was added after the experiment.
Phrases include all forms of tense;  phrases with \ddag\ include only past or perfect tense.
Phrases with \dag\ are included only when persons or objects appear as subjects.
The morpheme \textit{nai} is a lemma for
\textit{nu} and \textit{mase-n}.
}
\begin{tabular}{l}\\ 
a.\\
\hline
\hline
{\bf nigeru}\ddag/nigedasu\ddag/nukedasu\ddag/tousou\ddag/{\bf tabidatsu}\ddag/hassuru\ddag/\\
{\bf zurakaru}\ddag (run away)\\
e.g., kare wa \underline{nige ta} (he ran away)\\
\hline
{\bf nusumu}\ddag (stolen)\\
e.g., saihu ga \underline{nusuma re ta} (my wallet was stolen)\\
\hline
{\bf kieru}\dag/kiesaru\dag (disappear)\\
e.g., kare qa \underline{kie ta} (he disappeared) \\
\hline
{\bf nakunaru}\dag/
{\bf osimai}/{\bf mu-ni-modoru}/{\bf mu}/{\bf karappo}/{\bf aku}/utsuro/\\
{\bf monukenokara}/
\textit{hai-tte-nai}/\textit{noko-tte-i-nai}\dag/\textit{oka-re-te-i-nai}/\\
\textit{no-tte-i-nai} (nothing, empty)\\
e.g., subete ga \underline{mu ni modo tta} (everything returned to \textit{nothing})\\
\hline
{\bf yasumu}/\textit{suwa-tte-i-nai} (absent)\\
e.g., kare wa kyou \underline{yasun de iru} (he is absent today)\\
\hline
{\bf kaeru}\ddag (leave)\\
e.g., kare wa mou \underline{kae tta} (he has already left)\\
\hline
{\bf i-nai}/{\bf ga-nai}\dag/{\bf ha-nai}\dag/{\bf or-a-n} (be not here)\\
e.g., kare wa koko ni wa \underline{i nai} (he is not here)\\
\hline
{\bf mitsukara-nai}/\textit{miatara-nai} (not find)\\
e.g., teki ga \underline{mitsukara nai} (I can't find the enemy)\\
\hline
{\bf ko-nai}/\textit{modo-tte-ko-nai}/\textit{kae-tte-ko-nai} (not come)\\
e.g., kare wa mou kae tte \underline{ko nai} (he is not coming back)\\
\hline
\hline\\
b.\\
\hline
\hline
{\bf wasureru} (forget)\\
e.g., shorui wo hikitoru no wo \underline{wasure te i ta}\\
(I forgot to pick up the documents)\\
\hline
{\bf sibireru}/itamu/furueru (numb)\\
e.g., te ga \underline{sibirete iru} (my hands are numb)\\
\hline 
``$\dots$nai'' (not) other than the above list of existence negation\\
e.g., mie nai (cannot see)\\
\hline
\hline\\
\end{tabular}
\vspace{-1.5em}
\label{tab:phrase}
\end{table}

In
56.3\% of images with the annotation of existence negation (49 out of 87 images) and
33.3\% of images with the annotation of property negation (8 images out of 24 items), 
people significantly described negation, compared to baseline images (conditions 2 and 5). 
\begin{figure}[t]\center
\begin{tabular}{cc}
\includegraphics[scale=0.32]{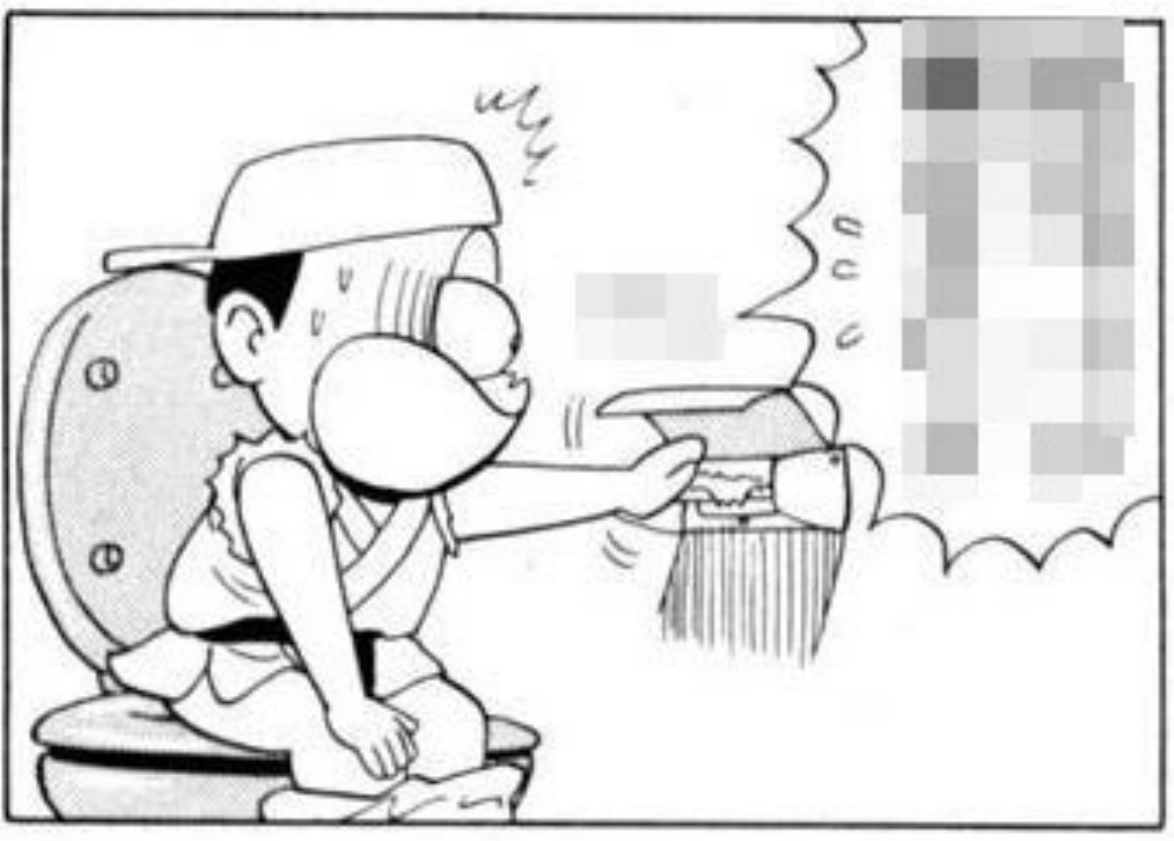} &
\includegraphics[scale=0.34]{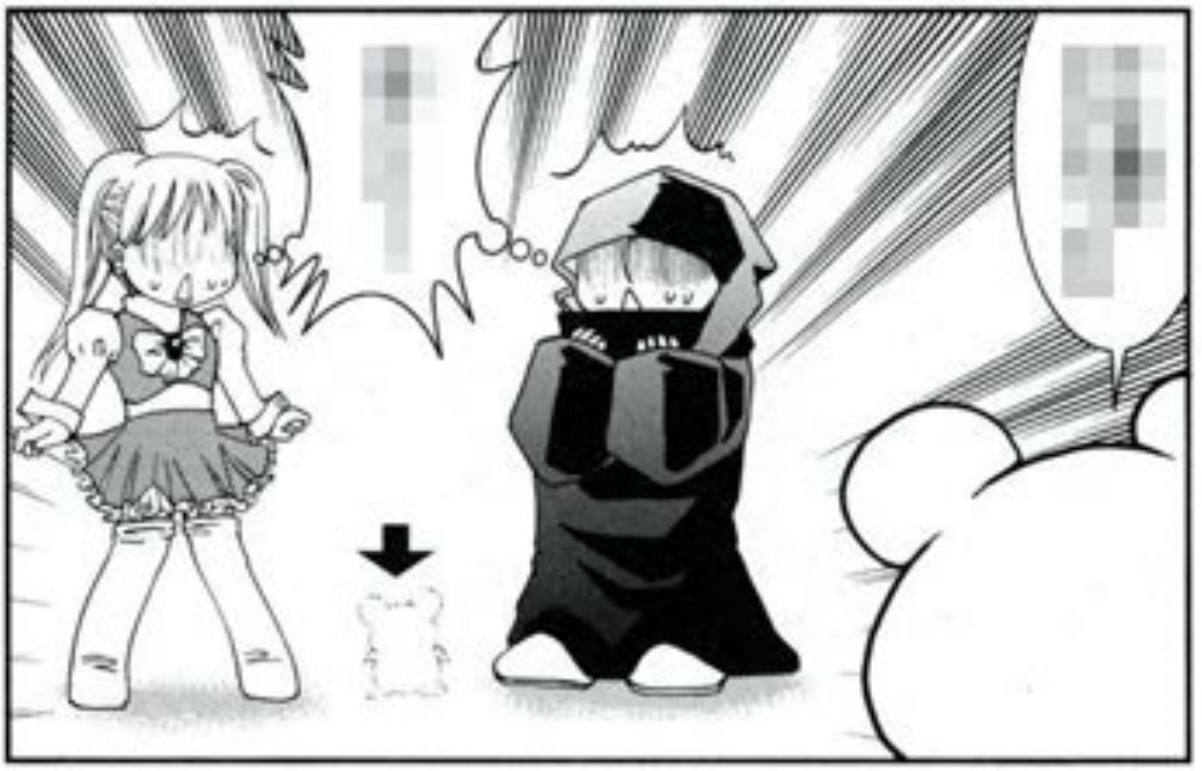} \\
(a) & (b)\\
\end{tabular}

\vspace{0.5em}

\begin{tabular}{cc}
\multicolumn{2}{c}{
\includegraphics[scale=0.39]{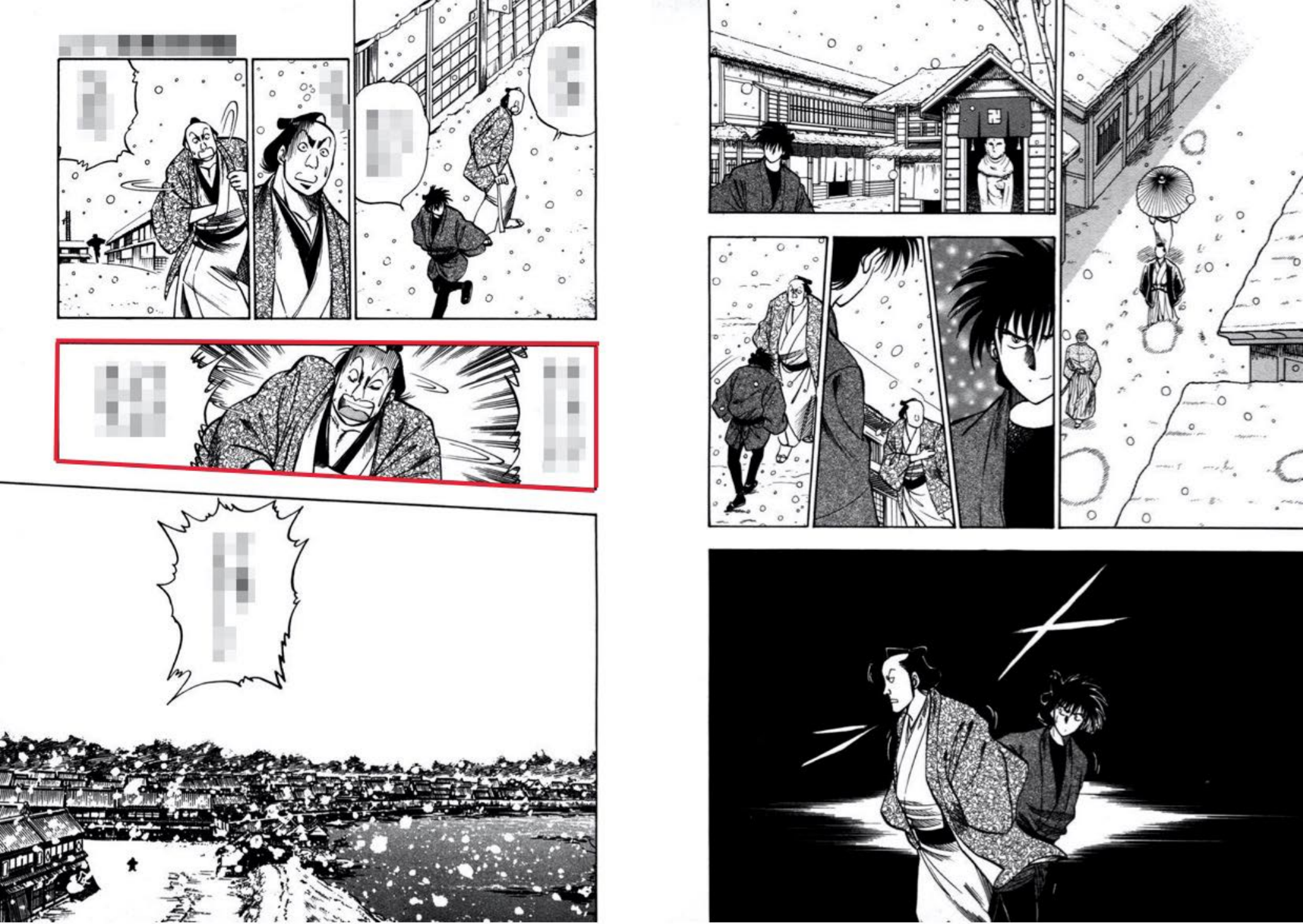}}\\
\multicolumn{2}{c}{(c)}\\
\end{tabular}

\caption{\footnotesize 
(a) Negation can be depicted without any aids: ``MoeruOnisan vol.19'', p.58{\copyright}Tadashi Sato / Shueisha; 
(b) Negation can be depicted with conventional devices: ``Akuhamu'', p.77 {\copyright}Satoshi Arai / Kodansha; 
(c) Negation can be depicted with sequences: ``BurariTessenTorimonocho'', pp.36-37 (the fourth panel of the left page is specified) {\copyright}Atsushi Sasaki / Shueisha}
\end{figure}

Detailed data of each image is presented in Appendix\footnote{\scriptsize \url{https://www.dropbox.com/s/niffpzg99p07dlv/app_cs21.pdf}}.
Seventeen images were able to depict negation without the need for sequences or conventional devices, i.e., by means of pure visual elements (16 images for existence negation; 1 image for property negation).
Fig.3a is an instance of this type (existence negation); 
$p <.01$ for condition 1 vs 2 and 
$p <.01$ for condition 4 vs 5.
Thirty-four images 
were able to depict negation under the constraint of 
using sequences (29 images for existence negation; 5 images for property negation).
Fig.3c is an instance of this type (existence negation); 
$p <.01$ for condition 1 vs 2 and 
$p >.1$ for condition 4 vs 5.
Two images were able to depict (existence) negation under the constraint of using conventional devices. 
Fig.3b is an instance of this type;
$p <.01$ for condition 1 vs 2,  
$p >.1$ for condition 3 vs 2, 
$p <.01$ for condition 4 vs 5,
$p >.1$ for condition 6 vs 5.
Three images were able to depict (existence) negation under the constraint of using either sequences or conventional devices (2 images for existence negation; 1 image for property negation). 
Fig.2 is an instance of this type (existence negation); 
$p <.01$ for condition 1 vs 2,  
$p <.1$ for condition 3 vs 2, 
$p <.01$ for condition 4 vs 5,
$p = .030$ for condition 6 vs 5.
One image was able to depict negation under the constraint of using both sequences and conventional devices.
``ChiisanaKoiNoMonogatari vol25'' (p.131) by Chikako Mitsuhashi is
an example of this type
(the image is included here because the permission to use has not been obtained yet); 
$p <.01$ for condition 1 vs 2,  
$p >.1$ for condition 3 vs 2, 
$p <.01$ for condition 4 vs 5,
$p >.1$ for condition 6 vs 5.

Given these empirical findings, 
we were able to offer a positive answer to the question of 
whether there are visual representations that can be recognized as expressing negation.
Some of the comic illustrations, 
especially 17 images (as in Fig.3a), 
could depict negation without the aid of sequence or conventional devices.

\section{Image classification task}

The analyses so far have resulted in the construction of the
\textsc{NegComic} dataset with fine-grained annotations,
which also revealed
the salient syntactic components expressing negation in illustrations.
The next step is to challenge the question: 
what distinguishes illustrations that represent negation from those that do not? 
To address the question, 
we analyzed machine learning (deep learning) performance and human performance 
in the task to classify illustrations into those containing negation or those not containing negation. 

Here our analysis focused on 
the 18 images that were judged as depicting negation without the need for sequence or conventional devices
(we added one image whose statistical significance was found 
at a reduced threshold of $10\%$
to the 17 images obtained in the experiment 1) and the corresponding negation-free images.

\subsection{Machine (deep) learning}
\subsubsection{Setup}
The procedure of analyses was as follows.  

\noindent
(1) {\bf Data augmentation}.
36 images were augmented into 648 by using the standard techniques such as contrast adjustments, smoothing, noise addition, and reverse turning in OpenCV\footnote{\url{https://github.com/opencv/opencv}}.  

\noindent
(2) {\bf Data split. }
The images were divided into 
training, validation, and test sets.
The three sets were completely independent.
The test set consisted of two images, negation (e.g., Fig.4a) and negation-free (e.g., Fig.4b); here the corresponding augmented images were removed from the analyses.  
\begin{figure}[t]\center
\begin{tabular}{cc}
\includegraphics[scale=0.25]{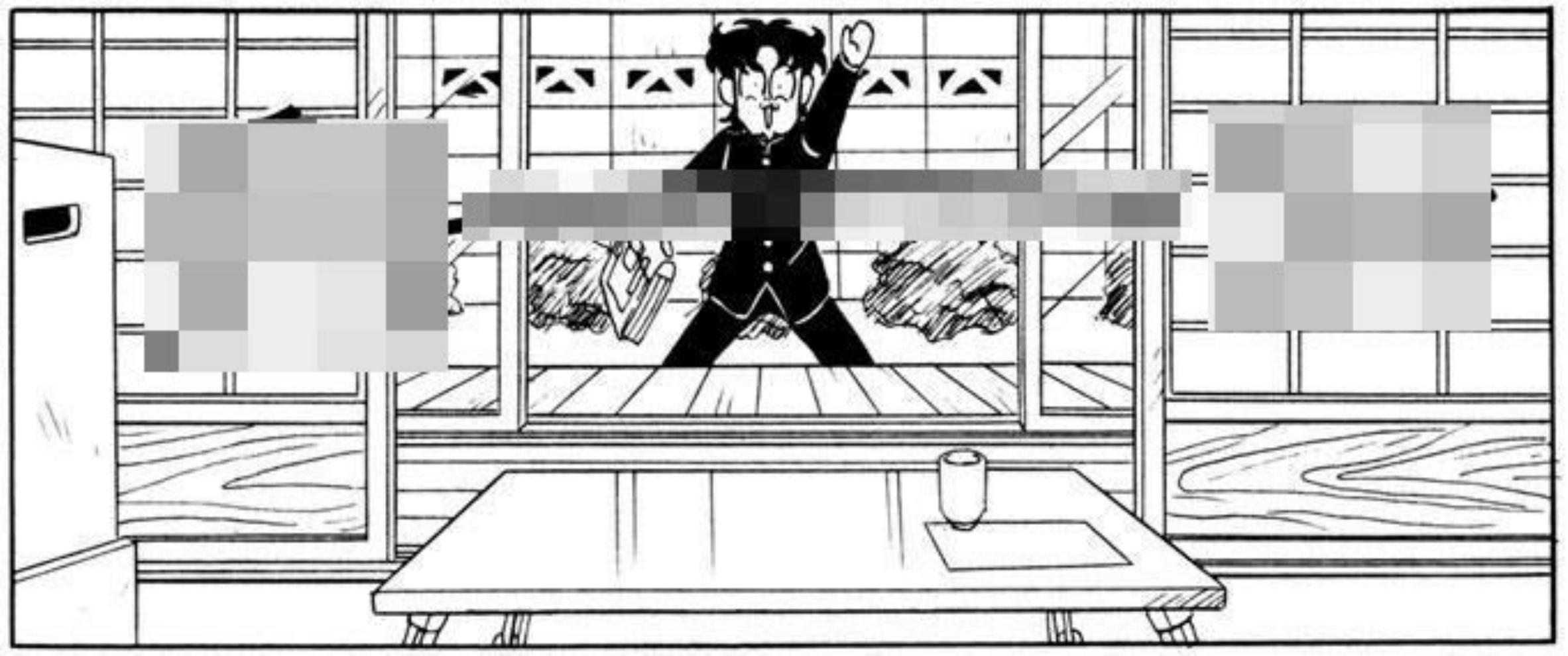} & 
\includegraphics[scale=0.21]{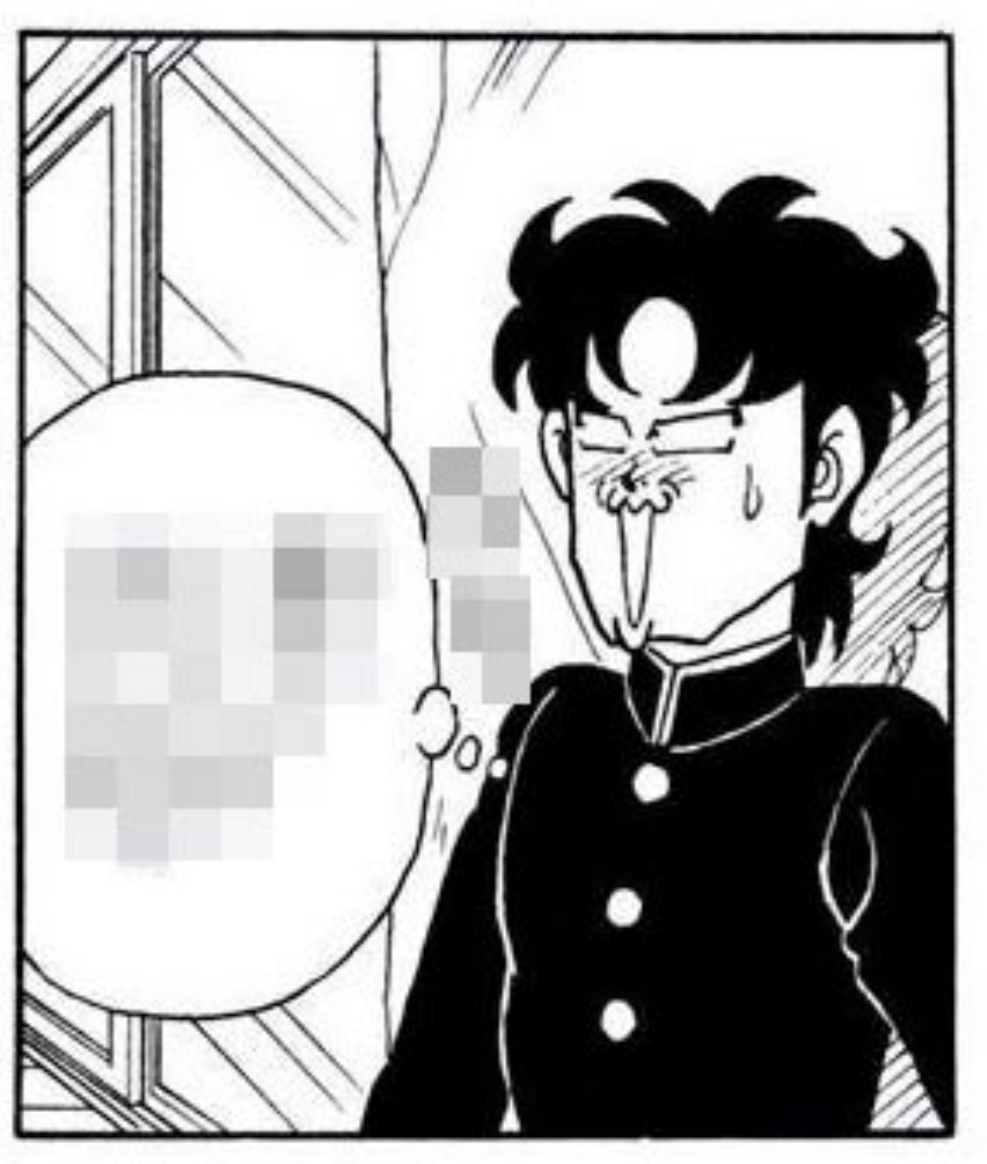}\\
(a) & (b)\\
\end{tabular}

\caption{\footnotesize 
Task example of classification tasks, ``HighschoolKimengumi vol.20'' pp138-139 {\copyright} Motoei Shinzawa / Shueisha; (a) negation type image,  (b) negation-free type image}
\end{figure}
Five hundred and four
images (18 original images) belonged to 2 classes (negation, negation-free) for learning and 
108 images (6 original images) belonged to 2 classes (negation, negation-free) for validation.
The assignment was randomly conducted and repeated 18 times; that is, 18 conditions were made.

\noindent
(3) {\bf Learning. }
We build a training model on a convolutional neural network (CNN) with a fine tuning technique using the pre-trained model of VGG16 (Simoyan \& Zisserman, 2015), in the environment of Python deep learning library Keras.
Key parameters are as follows:
we use sequential model, 
activation function for intermediate layer is Relu,
dropout rate is 0.5, 
activation function for output layer is Softmax,
VGG16 model weights for up to 14 layers, 
loss function is Crossentropy, 
optimizer is SGD,
batch size is 18, and epoch is 3.

\subsubsection{Test results}
Our CNN model showed that 
average $61.1\%$ of the 18 negation images were correctly classified as negation, as in Table 3.  
The rate of the negation-free images also was the same as that of the negation images (see the Appendix for each image result data in machine and human experiments).
\begin{table}[t]
\vspace{-1em}
\center
\caption{\footnotesize
Average accuracy of image classification tasks}
\vspace{-1em}
\begin{tabular}{lcc}\\ 
\hline
\hline
 & CNN & Human\\
\hline
negation & $61.1\%$ & $84.3\%$\\
negation-free & $61.1\%$ & $84.1\%$\\
\hline
\hline\\
\end{tabular}
\vspace{-1.5em}
\label{tab:classification}
\end{table}

\subsection{Human experiment}
In addition to the machine learning experiment, 
we examined the corresponding performance of ordinary people.

\subsubsection{Method}
Two hundred and five participants were recruited online; the basic setup was the same as in the experiment 1.  
The mean age of participants was 39.69 (SD = 9.66) with a range of 20-72 years.
The participants were asked to classify comic illustrations as those containing negation or those not containing negation. 
They were given the instruction on typical cases of negation:
``\textit{there is no \hole}'',
``\textit{\hole\ does not exist}'',
``\textit{\hole\ is not \hole}'',
``\textit{\hole\ disappeared}'',
``\textit{\hole\ is empty}'',
``\textit{\hole\ cannot do \hole}'',
``\textit{\hole\ does not move}''.
First,
they were asked to classify 18 images (half for negation, the other half for negation-free), which were randomly selected from 18 negation images and 18 negation-free images.
After answering the questions, 
they were given information to confirm if their answers were correct.
Then, this process was repeated using the same images.
The main test of image classification follows the second phase of
training. Four images (half for negation, the other half for negation-free) were randomly selected from 
the images other than the above 18 ones used for training.

\subsubsection{Test results}
In human participants, 
average $84.3\%$ in the 18 negation images were correctly classified as negation, as in Table \ref{tab:classification}.  
The rate of the negation-free images was $84.1\%$.

\subsection{Discussion}
The percentage of correct responses in the human negation classification task was high (84\%); the participants were able to distinguish negation images from negation-free images to some extent. 
Fig.~3(a) shows a case with a high percentage of correct responses. The percentage of correctly classified as negation was 100\% (21 out of 21 persons). While the caption of this image is \textit{There is no toilet paper},
the target of this description (\textit{toilet paper}) is not directly depicted in the image. 
This suggests that some background knowledge (``toilets usually have toilet paper'') is needed to recover the relevant information.
The use of background knowledge can be made for other images that had a high percentage of correct answers. 

The overall correct response rate (61\%) for the deep learning model is not much different from the 50\% chance level, suggesting that the model
struggled with the classification of negation images and negation-free images. Although the size of the training data is limited,
this in turn suggests that the difficulties of learning background knowledge can be a bottleneck in this task (cf. Bernardi et al., 2016; Garcia-Garcia et al., 2018).
The overall results of human and machine experiments using image classification tasks suggest that background knowledge can play an important role in recognizing negation.

\section{Conclusion and future works}
The data collected in image captioning tasks 
showed that
some comic illustration images  
can depict negation even without the aid of sequence or conventional devices.
This type of comic image was subjected to further human and machine experiments on image classification.
The performance results shed light on the view that 
some comic images evoke the knowledge presupposed
when recognizing negation
and thus can depict negation with purely visual elements.
This is substantially different from linguistic representations that express negation while referring to the situation or the object to be negated, such as ``the train is not coming'' as mentioned in Introduction.
To recognize the negation from the image,
it is crucial to infer what is being negated
from a piece of background knowledge that is not directly depicted
in a given image.

In the next stage, more research will be directed to providing a mechanism to focus on specific background knowledge
and to clarifying the semantic conditions for context-dependent knowledge.
Although the experiments suggested some negative results on
the machine learning model,
we do not intend to argue the limitations of machine learning
approaches in general.
Rather, we aim to shed light on the interaction between cognitive science and machine learning,
which could contribute to achieving
the overall goal in AI, i.e., to build a machine that thinks and understands images like humans (cf.\,Lake et al., 2017).
Currently, image recognition based on deep learning is often said to rival or surpass human accuracy (e.g., Rawat \& Wang, 2017).
However, our experiments suggest that the accuracy of
a deep learning model
is achieved in a different way from the actual human cognitive process that crucially involves knowledge and common sense.

In addition to negation, we plan to expand our research to include disjunction (\textit{or}), conditionals (\textit{if... then}),
and quantifiers (\textit{all}, \textit{most}, \textit{some}).
The ability to manipulate such logical information seems to be an inherent ability of humans, who manipulate languages. It would be interesting to examine from empirically collected data
whether it is possible to handle such logical information as a visual representation.

\section*{Acknowledgments}
This work was supported by JSPS KAKENHI Grant Number JP20K12782 to the first author. 

\bibliographystyle{apacite}

\bibliography{CogSci_Template}

\end{document}